\documentclass{Interspeech}

\interspeechcameraready

\title{LID Models are Actually Accent Classifiers: Implications and Solutions for LID on Accented Speech}

\author[affiliation={1},equalcontribution]{Niyati}{Bafna}
\author[affiliation={1,2},equalcontribution]{Matthew}{Wiesner}

\affiliation[nocounter]{$^{1}$CLSP and $^{2}$HLTCOE}{Johns Hopkins University}{USA}

\email{nbafna1@jhu.edu, wiesner@jhu.edu}

\newcommand{\et}{\texttt{ECAPA-TDNN}}
\newcommand{\phoneseq}{\texttt{phoneseqs}}
\newcommand{\duseq}{\texttt{duseqs}}
\newcommand{\etps}{\texttt{ET+phoneseqs}}
\newcommand{\etpstrain}{\texttt{ET+phoneseqs-train}}
\newcommand{\etdutrain}{\texttt{ET+duseqs-train}}
\newcommand{\etduembedtrain}{\texttt{ET+duseqsembed-train}}

\usepackage{comment}
\usepackage{hyperref}
\usepackage{multirow} 
\usepackage{adjustbox}

\usepackage{colortbl}
\usepackage{xcolor}
\usepackage{graphicx}
\usepackage{multirow}
\usepackage{booktabs}

\definecolor{white}{RGB}{255, 255, 255}
\definecolor{lightestgreen}{RGB}{250, 255, 240}
\definecolor{lightergreen}{RGB}{225, 240, 205}
\definecolor{lightgreen}{RGB}{200, 225, 170}
\definecolor{midgreen1}{RGB}{175, 210, 135}
\definecolor{midgreen2}{RGB}{150, 195, 100}
\definecolor{darkgreen}{RGB}{125, 180, 65}
\definecolor{darkergreen}{RGB}{100, 165, 30}
\definecolor{darkestgreen}{RGB}{75, 150, 0}

\definecolor{lighterred}{RGB}{255, 220, 220}
\definecolor{lightred}{RGB}{235, 180, 180}
\definecolor{red}{RGB}{215, 140, 140}
\definecolor{darkred}{RGB}{195, 100, 100}
\definecolor{darkerred}{RGB}{175, 60, 60}

\begin{document}

\maketitle

\begin{abstract}

Prior research indicates that LID model performance significantly declines on accented speech; however, the specific causes, extent, and characterization of these errors remain under-explored. (i) We identify a common failure mode on accented speech whereby LID systems often misclassify L2 accented speech as the speaker's native language or a related language.  (ii) We present evidence suggesting that state-of-the-art models are invariant to permutations of short spans of speech, implying they classify on the basis of short phonotactic features indicative of accent rather than language. Our analysis reveals a simple method to enhance model robustness to accents through input chunking. 
(iii) We present an approach that integrates sequence-level information into our model without relying on monolingual ASR systems; this reduces accent-language confusion and significantly enhances performance on accented speech while maintaining comparable results on standard LID.\footnote{We release our code at \url{https://github.com/niyatibafna/mitigating-accent-bias-in-lid/}.}
\end{abstract}

\section{Introduction}
\label{sec:introduction}
Spoken language identification (LID) is vital for speech processing pipelines, but often fails to generalize across diverse accents. Accent variation arises for many reasons: in national and global link languages such as English, Spanish, and Swahili, speakers color their pronunciation with the phonology of local substrate languages or their L1 languages \cite{flege1995second}.
For example, Indian English speakers may approximate the fricative $/\theta/$ in English with a dental $/d/$, since most Indian language phone inventories do not contain the former.
Other kinds of L1 accent variation may not be easily described by a currently spoken substrate or L1 language phonology, e.g., Latin American Spanish accents. While all speakers have an accent \cite{markl23interspeech}, we focus specifically on how LID models handle accents influenced by local substrate or L1 substrate phonology --  L2-accents for short.\footnote{For bilingual speakers, or L1 speakers who share similar accent features as a group of L2 speakers we still call this accent an L2-accent.}

We aim to broadly characterize the mode of errors made by modern LID approaches on multiple L2-accents in 3 languages, to explain why some models fail to generalize to L2-accents, and finally, informed by our analyses, to improve LID robustness to L2-accented speech.
To this end, we first study the error patterns of the ECAPA-TDNN LID model \cite{valk2021slt,desplanques2020ecapa} on a variety of L2-accents.
We find that accent-language confusion, i.e. the mis-recognition of L2-accented speech as the L1 substrate or a related language, is a significant contributor to this degradation. For example, Catalan-accented English is often identified as Catalan, Filipino-accented speech is identified as Tagalog and Cebuano, and so on. 

Accent-language confusion in an LID model indicates that the model functions as an accent ID model and simply relies on the strong correlation between accent and language to make language predictions. We hypothesize that this occurs when models use phoneme inventories or short phonotactic features characteristic of accent to make predictions, instead of lexical or syntactic features that may more generally characterize language. 
We probe LID models on adversarially constructed inputs designed to help characterize invariance to block permutations in order to help explain errors on L2-accented speech. 

Current commonly used LID models, such as the ECAPA-TDNN \cite{desplanques2020ecapa}, or pooled classification of self-supervised (SSL) representations of speech,  borrow from techniques used in speaker ID. These models are not explicitly trained to capture sequence-level information, and involve pooling operations which inherently treat input sequences as exchangeable. For these reasons, they may be uniquely vulnerable to mis-classifying L2-accents. Formerly, phonotactic models were commonly used for LID \cite{li2005phonotactic, 479741, 4013519, matejka2005phonotactic, d2012phonotactic}, and some recent work has looked at fusing these approaches to improve LID. Our analyses indicate that one reason for the success of these fusion approaches is that they are robust to L2-accents.

Finally, we explicitly incorporate sequence-level views of the data into our models to improve accent-robustness.
We explore two methods of extracting coarse sequence-level information from the input signal, and train transformer-based LID classifiers that take these as sequence inputs: a) phonetic transcripts (\phoneseq), and b) clustered SSL representations (\duseq).
The latter is inspired by \cite{lakhotia2021generative,cormacenglishetal2022domain,millet2022self}, who show that discrete SSL units largely encode phonemic information.
We show these models display very little accent-language confusion, displaying largely accent-agnostic error patterns for English. 
This lends support to the claim that sequence-level features contribute to accent-robustness for LID.

However, these models show poor overall LID performance as compared to acoustics-based SOTA model.
This indicates that acoustic and phonetic representations provide complementary information and can therefore be beneficially combined.
We explore combinations of phonetics-based and acoustics-based views of the data, including fusion of model output distributions (\etps), as well as using frozen \et{} representations with trainable phonetics-based modules (\etpstrain, \etdutrain, and \etduembedtrain).
Our best-performing model, \etpstrain{}, shows large gains on L2-accents (up to $+34$ LID points for English L2 accents), while suffering minimally on LID for mainstream accents.

Our work sheds light on a prevalent mode of and reasons for the failure of SOTA LID systems on accented speech,  shows that sequence-view-based LID models are less susceptible to this problem, and demonstrates the benefits of incorporating local and global views of the data on accented speech.

\section{Related Work}
\label{sec:related_work}
The degradation of commonly used LID and ASR systems on accented speech is well documented \cite{najafian2020automatic,stylesinvestigating,fengetal2021quantifying,sanabriaetal2023edinburgh,shietal2021accented}, although largely only for English dialects and accents. We extend this analysis to a broader set of accents and languages than previously studied for LID systems.
Some previous work has investigated using ASR outputs to aid LID systems on accented speech: \cite{chandaketal2020streaming} use ASR hypotheses for multiple monolingual ASR systems as input to an LID system, showing improvements on accented and code-switched speech.
Other studies have also explored improving LID by using phonetic information; \cite{shahin2023improving} show that fine-tuning \texttt{wav2vec2} on articulatory feature detection improves its performance on English/Mandarin LID. \cite{liu2022pho} proposed PHO-LID, which uses an additional SSL-based contrastive phoneme segmentation objective to inject phonotactic information into the model and reduces confusions between nearby languages. Our modeling  approach, while similar, is somewhat simpler and we primarily use it to analyze how phonotactic information improves performance on L2-accents.

Our work is similar to \cite{kukkalumae2022improving}, which uses a naïve Bayes char-gram model monolingual-ASR transcripts to inform LID predictions using fusion with the SOTA model predictions. To our knowledge, ours is the first work to characterize accent-language confusion as a major cause of failure for LID on accented speech, and link it with the block permutation invariance of LID models.
Our model uses a language-agnostic phonetic transcriber instead of relying on ASR systems, and is trained with access to SOTA model representations, which outperforms a shallow fusion-based approach.

\section{Datasets} 
\label{sec:datasets}
See \autoref{tab:datasets} for a summary of the datasets we used.

\textbf{General domain training} -- The VoxLingua-107 (VL-107) \cite{valk2021slt} dataset is used to train multi-domain baseline models since it has broad coverage of languages, accents, and acoustic channels and we are interested in evaluating LID models on typical data found ``in the wild.'' VL-107 contains a total of $6628$ hours of speech from YouTube covering  $107$ languages. We assume the speech consists mainly of common, L1 accents, though no ground-truth accent labels are provided.
In order to remove the effect of sequence length from all of our analysis on models that we train, we chunk both our training and evaluation data into $6$ second utterances, and treat each such segment as a different sample for evaluation purposes.

\textbf{Common L1 (mainstream) accents} -- We want to verify that improvements from our approach on accented speech do not come at the cost of degraded performance on speech in mainstream accents or other languages. We report general-purpose LID performance on the FLEURS \cite{conneau2023fleurs} test set, which provides L1 speech from a relatively clean domain.

\textbf{L2 accents in English} -- For evaluation on English accented speech, we use: 1) \emph{The Edinburgh International Accents of English Corpus} (EdAcc) ~\cite{sanabriaetal2023edinburgh}, containing English accented conversational speech. 2) \emph{CommonVoice} (CV): a subset containing English accented speech from v1.0.
3) \emph{The Speech Accent Archive} (SAA) \cite{saa}, as an additional set of recorded accents with the special property that all of the speakers say exactly the same sentence. The version we downloaded contains 16.5 hrs of speech from 2138 speakers, 200 unique accents, with 68 accents containing at least 5 examples of speech.

\textbf{L2 accents in German / French} -- We filtered CommonVoice v13.0 \cite{commonvoice2020} for mainstream and L2 accented data in German and French (e.g. Polish-accented German). We collected 1.3 hrs of speech from $10$ L2-German and $6$ L2-French accents.

The self-reported accents in EdAcc are not standardized.
Therefore, we manually grouped them into $35$ L2-accent categories totaling $19$h, excluding L1 accents (e.g. ``Australia'').
CommonVoice accent reporting is relatively normalized, containing $3.9$h with $6$ L2 English accents, although with varying granularity, e.g., there is a single ``African'' accent. The SAA is primarily used for analysis where it was important to have controlled for the spoken content of the speech.

\begin{table}[ht]
\centering
\caption{Summary of datasets used in the study.}
\adjustbox{max width=\linewidth}{
\begin{tabular}{|c|c|c|c|c|}
\toprule
Use & Dataset & \# hr & \# Accents & \# Langs \\
\toprule
Train & VL-107 \cite{valk2021slt} & 6.6k h &  - & 107 \\
\hline
Test L1 & FLEURS Test \cite{conneau2023fleurs} & 283 h & - & 102 \\
\hline
\multirow{3}{*}{Test L2 en} & EdAcc \cite{sanabriaetal2023edinburgh} & 19 h & 35 & 1 \\
& CV v1 & 3.9 h & 8 & 1 \\
& SAA \cite{saa} & 16.5 h & 200 & 1 \\
\hline
Test L2 fr, de& CV 13.0 & 1.3 h & 10 de, 6 fr & 2 \\
\hline
\end{tabular}
}
\label{tab:datasets}
\end{table}

\section{Models} 
\label{sec:models}

\begin{figure}[!ht]
    \centering
    \includegraphics[width=0.8\linewidth]{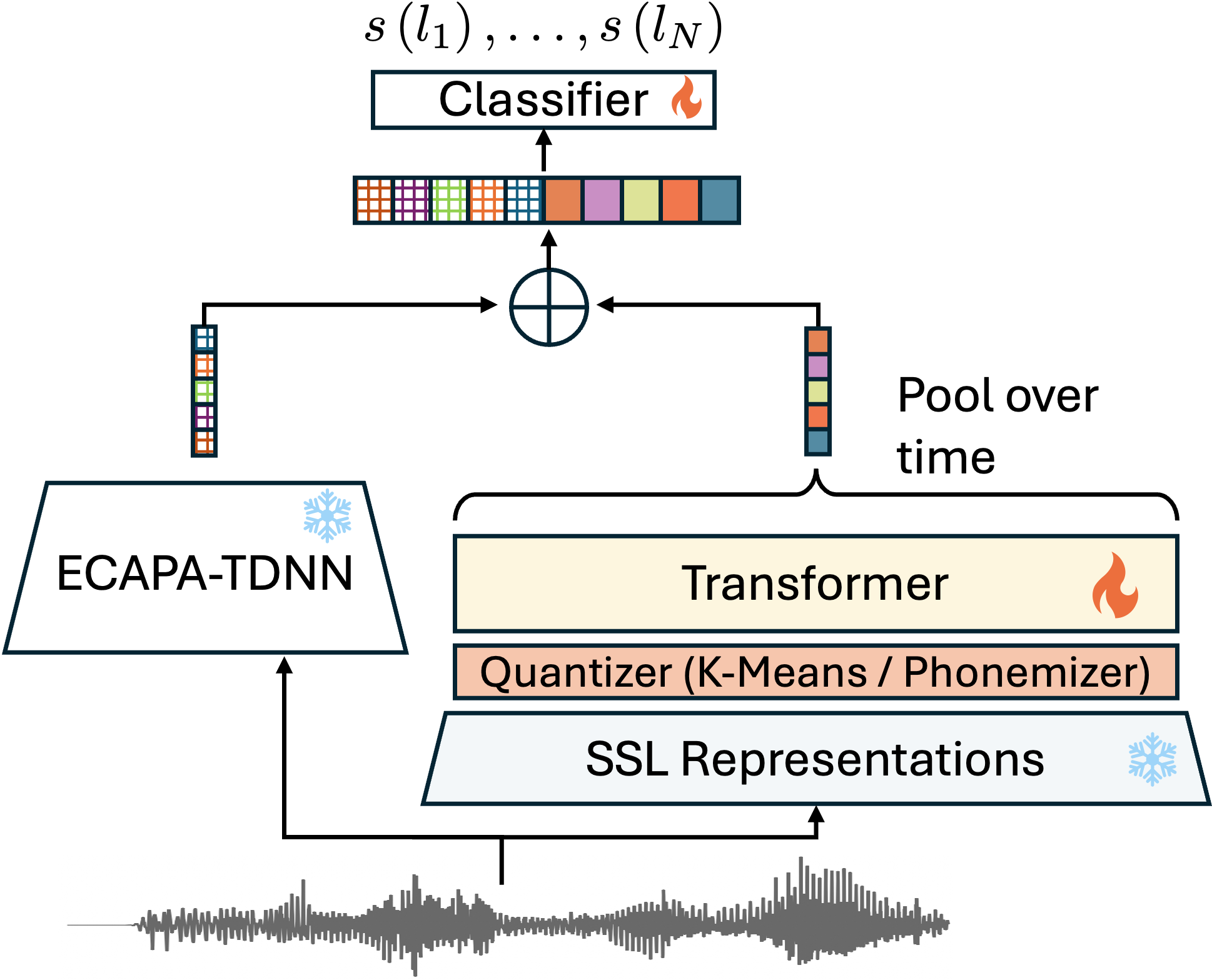}
    \caption{The depicted model augments the ECAPA-TDNN representation with one produced by passing a quantized representation of speech into a learned transformer model. \etpstrain{} uses a phonetic transcript of the audio, while \etduembedtrain{} quantizes SSL representations with K-means clustering. For \etdutrain{}, the transformer embedding layer is initialized as the K-means centroids. The classifier produces scores $s\left(l_i\right)$ for each language, $l_i$, among $N$ possibilities.}
    \label{fig:model}
    \vspace{-2mm}
\end{figure}

\subsection{Baselines and Analysis}
We use the \et{} model (21M parameters), trained for LID on VoxLingua-107, as our baseline, and conduct our error pattern and context window analyses on this model.
We also repeat the latter on two top-performing LID models: 1) the MMS model (1B parameters), trained on FLEURS, 2) the GEO model from \cite{foleyetal2024geolocating}, which is built off of the MMS SSL model and trained for speech geolocation on 3k hr of speech. Fine-tuning the resulting model for LID outperformed the MMS model.

\subsection{Using phone transcripts}
We use \texttt{wav2vec2-xlsr-53-espeak-cv-ft} \cite{xu2022simple} to generate phonetic transcriptions of the text. 
Our \phoneseq-component takes the phoneme sequences as input, treating each phone as a separate token. 
It has an embedding layer with dimension $256$, followed by $8$ transformer layers, with attention dimension $128$ and $8$ attention heads, and a linear classification layer ($1.2M$ parameters in total).
\etpstrain{} (depicted in \autoref{fig:model}) concatenates representations from \et{} (frozen) to the \phoneseq-module representations before the classification layer during training.
We also provide results using only \phoneseq{}, as well as a fusion-based model (\etps) that averages output probability distributions of \et{} and \phoneseq{}.

\subsection{Using discrete SSL units}
Our \duseq{} (discrete-unit sequence) model uses discretized \texttt{wav2vec2-base} representations in lieu of phone sequences.
We obtain representations from the 8th layer of \texttt{wav2vec2-base} (as per \cite{lakhotia2021generative}), pool over 100ms segments, (rough duration of uttered phones), and train KMeans clustering over the resulting representations obtained over all training languages, using $1000$ clusters.
The input to the \duseq{} model is therefore a sequence of centroids clusters.
\etdutrain{} (depicted in \autoref{fig:model}) uses \et{} representations analogously to \etpstrain{}, with a \duseq-component using centroid representations from the KMeans clustering as embeddings ($768$-dim), $4$ transformer layers with attention dimension $128$ and $8$ attention heads, followed by a linear classification layer ($0.6M$ parameters in total).
We further train \etduembedtrain, which learns \texttt{256}-dimensional embeddings for the centroid clusters from scratch during training, and provide an ablation using only \duseq.

All models are trained on VL-107, on a single GPU with learning rate $1e-4$; \phoneseq{} and \duseq{} for $20$ epochs, and the \texttt{ET+*} models for $10$ epochs.

\section{Error profiles on accented speech}
\label{sec:error_profile_analysis}

Confirming previous work, we find a significant disparity in model performance on mainstream high-resource accents (\texttt{uk, us, canada}) and L2 accents in English.
\et{} has a mean accuracy of $87.6\%$ and $73\%$ for mainstream accents in CommonVoice and EdAcc respectively, but degrades to a mean accuracy of $55.8\%$ and $57\%$ respectively on L2 accents.
We also show this problem for German and French L2 accents: \et{} performs with mean $93.4\%$ and $97.5\%$ accuracy on German and French mainstream accents respectively, but degrades to $61.3\%$ and $80.1\%$ accuracy respectively for L2 accents in these languages.

\subsection{Accent-language confusion}
\label{sec:acc_lang_confusion}
\begin{figure}[!ht]
    \centering
    \hspace*{-8mm}
    \includegraphics[scale=0.48]{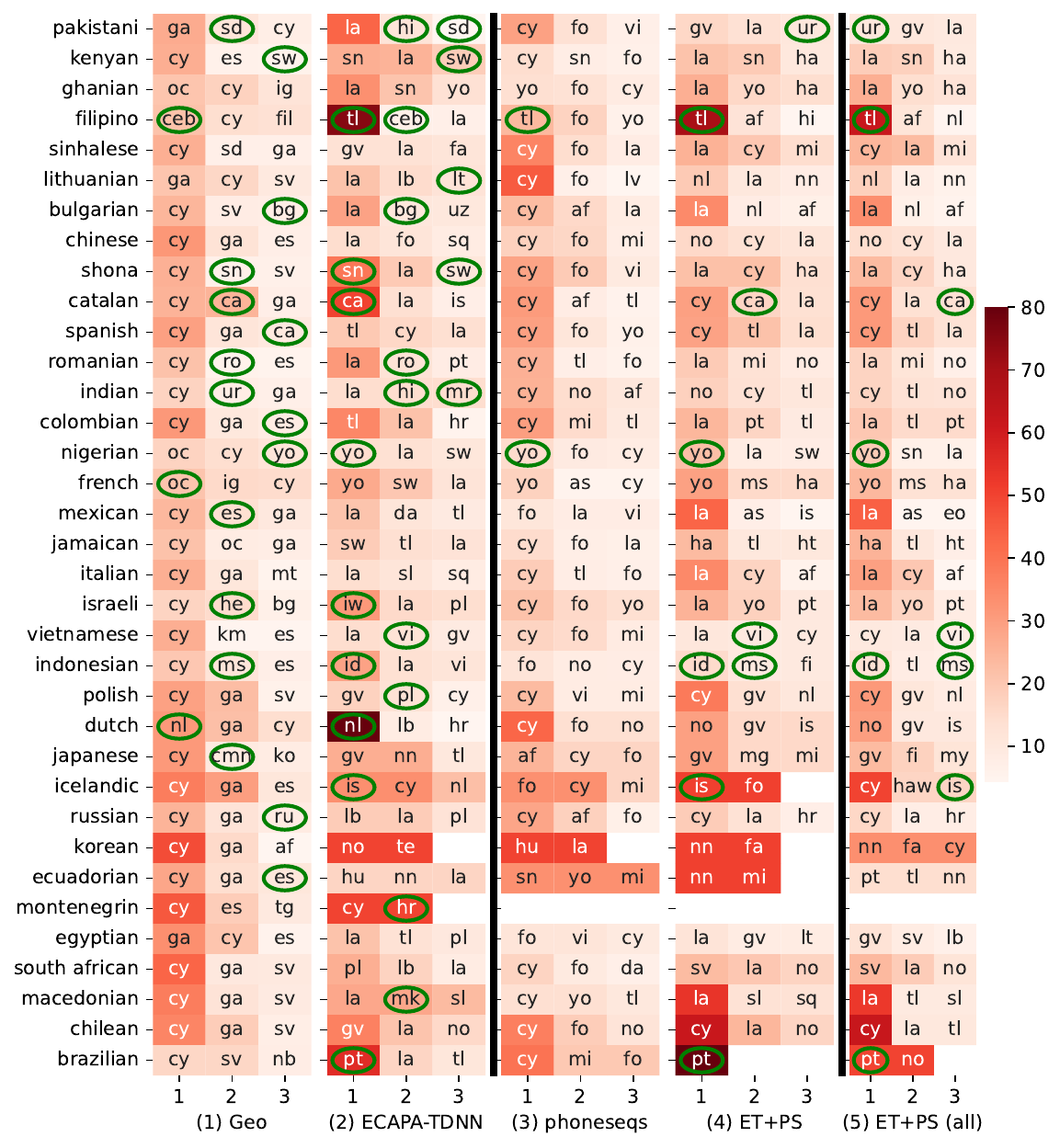}
    \caption{Error profiles for SOTA (1, 2) and our best-performing \etpstrain{} model (5) on EdAcc accents, showing the top 3 languages that each accent was mis-classified as, as well as associated percentage of total error. (3) and (4) show the error profile for \phoneseq{} and \etpstrain{} on samples where \et{} made mistakes.  Errors indicative of accent-language confusion are highlighted in green.}
    \label{fig:confusions}
    \vspace{-4mm}
\end{figure}

To test our first hypothesis -- that a common failure mode of LID system is due to mis-classification with a speaker's L1 language or a related language -- we study the confusions of \et{} on L2-accented data.
In \autoref{fig:confusions}, we demonstrate accent-language confusion in SOTA LID models, as well as its mitigation in our \phoneseq-based models.
For each accent, we examine the top 3 predicted languages for misclassified examples of speech as well as the associated percentage of total error for \et{} and the GEO model. 
We see that accent-language confusion often constitutes a significant portion of the error for \et{}.
For example, the mis-classification of Dutch-accented English as Dutch and Brazilian-accented English as Portuguese constituting 82.6\% and 50\% of total error respectively.
We also observe the same trend for \et{} on the German and French speech; e.g. the mis-classification of Polish-accented German as Polish comprises $33.3\%$ of total error.
The error profiles of \phoneseq{} and \etpstrain{} on the same set of samples show a clear contrast, most visible for \phoneseq{}, with much lower accent-language confusion, and similar errors largely regardless of accent. 
This suggests that the \phoneseq-component captures an alternate view of the data, useful for combating accent-language confusion.

\begin{table*}[ht]
\centering
\caption{``Mean'' shows the bootstrap resampling average over speakers with $95\%$ confidence intervals in parentheses. ``Macro'' shows the macro-averages with std. dev., computed over languages for FLEURS, and over accents for all other datasets. All results were significant as compared against the baseline \et{} system using a McNemar test with $p<0.05$.}
\adjustbox{max width=\linewidth}{
\begin{tabular}{l|cc|cc|cc|cc }
\toprule
 & \multicolumn{2}{c|}{\textbf{FLEURS}} & \multicolumn{2}{c}{\textbf{EdAcc (L2 en)}} & \multicolumn{2}{c}{\textbf{CommonVoice (L2 en)}} & \multicolumn{2}{c}{\textbf{CommonVoice (L2 non-en)}} \\
 \midrule
 & Mean & Macro & Mean & Macro & Mean & Macro & Mean & Macro \\
\midrule
\et & 89.3 (89.0,89.6)  & 89.5±17.2 & 47.9 (40.1,56.2)  & 55.8±26.8 & 34.5 (23.7,48.4)  & 57.0±24.6 & 63.8 (54.2,73.1)  & 68.4±22.2 \\
\etpstrain & 86.6 (86.1,87.0)  & 86.4±18.2 & \textbf{57.4 (48.9,65.6)}  & \textbf{64.0±25.4} & \textbf{68.9 (61.2,76.1)}  & \textbf{81.6±10.7} & \textbf{73.0 (63.3,81.0)}  & \textbf{76.0±13.9} \\
\midrule
\etps & \textbf{89.5 (89.2,89.9)}  & \textbf{89.5±17.8} & 52.2 (43.9,60.5)  & 59.8±26.5 & 46.4 (37.3,57.2)  & 69.1±18.6 & 66.8 (56.6,75.7)  & 73.6±20.3 \\
\phoneseq & 52.9 (52.1,53.7)  & 52.5±22.7 & 37.3 (30.5,44.2)  & 45.2±22.8 & 47.3 (40.9,54.8)  & 67.3±13.6 & 48.4 (40.6,56.0)  & 51.6±14.9 \\
\duseq & 49.6 (48.9,50.3)  & 49.8±18.3 & 42.6 (37.3,48.0)  & 48.6±17.8 & 48.3 (39.4,56.7)  & 66.3±13.5 & 48.1 (40.9,55.2)  & 48.2±19.2 \\
\midrule
\etdutrain & 84.7 (84.3,85.1)  & 84.9±18.9 & 50.7 (43.0,58.1)  & 58.3±24.4 & 48.1 (39.3,58.0)  & 67.0±15.6 & 68.6 (59.8,76.6)  & 70.0±22.7 \\
\etduembedtrain & 84.2 (83.8,84.7)  & 84.2±20.2 & 53.4 (45.9,60.9)  & 60.7±23.7 & 51.5 (43.5,60.7)  & 67.5±13.3 & 63.7 (54.8,71.3)  & 65.6±22.8 \\
\bottomrule
\end{tabular}
}
\label{tab:approach_results}
\vspace{-3mm}
\end{table*}

\begin{table}[ht]
    \centering
    \renewcommand{\arraystretch}{1.2}
    \newcommand{\best}[1]{\textbf{#1}} %
    \aboverulesep=0ex %
    \belowrulesep=0ex %
    \caption{The relative degradation (\%) in performance when the input audio is block-reversed. Colors range from light to dark red, with darker colors indicating that the model \textbf{does not} treat sequences of chunks of the corresponding size as exchangeable.}
    \adjustbox{max width=0.95\linewidth}{
    \begin{tabular}{l|l|ccccc}
        \toprule
        \multicolumn{2}{c|}{}&  \multicolumn{5}{c}{\bf{Chunk Size} (s)} \\
        \midrule
        Accent &  Model & 0.25 & 0.5 & 1 & 2 & 4 \\
        \midrule
        \multirow{3
    }{*}{en\_us} 
          & ECAPA & \cellcolor{lighterred}-2.7 & \cellcolor{white}-0.7 & \cellcolor{white}-0.5 & \cellcolor{white}0.0 & \cellcolor{white}-0.2\\
        & MMS & \cellcolor{white}-1.7 & \cellcolor{white}-0.5 & \cellcolor{white}-0.5 & \cellcolor{white}-0.2 & \cellcolor{white}-0.2\\
        & GEO & \cellcolor{lightred}-6.7 & \cellcolor{white}0.0 & \cellcolor{white}0.0 & \cellcolor{white}-0.2 & \cellcolor{white}0.0\\
        \midrule
        \multirow{3
    }{*}{other} 
        & ECAPA & \cellcolor{lighterred}-4.4 & \cellcolor{white}-1.9 & \cellcolor{white}0.7 & \cellcolor{white}-0.2 & \cellcolor{white}0.4\\
        & MMS & \cellcolor{lightred}-5.6 & \cellcolor{white}-0.9 & \cellcolor{white}1.1 & \cellcolor{white}-0.4 & \cellcolor{white}0.0 \\
        & GEO &\cellcolor{darkerred}\textcolor{white}{-37.0} & \cellcolor{darkred}\textcolor{white}{-15.8} & \cellcolor{lightred}-6.0 & \cellcolor{lighterred}-3.2 & \cellcolor{lighterred}-2.4\\
        \bottomrule
    \end{tabular}
    }
    \label{tab:rev_results}
    \vspace{-3mm}
\end{table}

\begin{table}[ht]
    \centering
    \renewcommand{\arraystretch}{1.2}
    \newcommand{\best}[1]{\textbf{#1}} %
    \aboverulesep=0ex %
    \belowrulesep=0ex %
    \caption{Accuracy results for ECAPA-TDNN, MMS, and GEO models on EdAcc and SAA datasets where segments are chunked into varying window sizes and predictions are aggregated across the chunks by majority vote. The highest accuracy achieved for each accent category in each dataset is \textbf{bolded}. Colors range from light to dark green, with darker indicating better performance.}
    \adjustbox{max width=\linewidth}{
    \begin{tabular}{l|l|l|ccccc}
        \toprule
        \multicolumn{3}{c|}{}&  \multicolumn{5}{c}{\bf{Window Size} (s)} \\
        \midrule
        Accent & Dataset & Model &  0.5 & 1 & 2 & 4 & 8 \\
        \midrule
        \multirow{6
    }{*}{en\_us} 
        & \multirow{3}{*}{EdAcc}  & ECAPA &  \cellcolor{midgreen2}92.3 & \cellcolor{darkestgreen}100.0 & \cellcolor{darkestgreen}100.0 & \cellcolor{darkestgreen}100.0 & \cellcolor{darkestgreen}\bf{100.0}\\
        & & MMS & \cellcolor{lightgreen}73.0 & \cellcolor{darkergreen}96.0 & \cellcolor{darkergreen}96.0 & \cellcolor{darkergreen}96.0 & \cellcolor{darkestgreen}\bf{100.0}\\
        & & GEO &  \cellcolor{midgreen1}88.5 & \cellcolor{darkergreen}96.2 & \cellcolor{darkestgreen}100.0 & \cellcolor{darkestgreen}100.0 & \cellcolor{darkestgreen}\bf{100.0}\\
        \cmidrule{2-8}
        & \multirow{3}{*}{SAA}  & ECAPA & \cellcolor{midgreen1}84.0 & \cellcolor{darkergreen}96.4 & \cellcolor{darkestgreen}99.0 & \cellcolor{darkestgreen}98.8 & \cellcolor{darkestgreen}99.0\\
        & & MMS & \cellcolor{midgreen1}88.5 & \cellcolor{darkestgreen}99.3 & \cellcolor{darkestgreen}99.5 & \cellcolor{darkestgreen}99.8 & \cellcolor{darkestgreen}\bf{100.0}\\
        & & GEO &  \cellcolor{darkestgreen}98.8 & \cellcolor{darkestgreen}100.0 & \cellcolor{darkestgreen}100.0 & \cellcolor{darkestgreen}100.0 & \cellcolor{darkestgreen}\bf{100.0} \\
        \midrule
        \multirow{6
    }{*}{other} 
        & \multirow{3}{*}{EdAcc}  & ECAPA & \cellcolor{lightestgreen}23.5 & \cellcolor{lightestgreen}39.8 & \cellcolor{lightergreen}51.5 & \cellcolor{lightergreen}60.0 & \cellcolor{lightergreen}65.1\\
        & & MMS  & \cellcolor{lightestgreen}21.0 & \cellcolor{lightergreen}51.0 & \cellcolor{lightergreen}63.0 & \cellcolor{lightergreen}65.0 & \cellcolor{lightergreen}66.0 \\
        & & GEO & \cellcolor{lightergreen}55.9 & \cellcolor{lightgreen}81.3 & \cellcolor{midgreen1}\bf{84.0} & \cellcolor{lightgreen}80.0 & \cellcolor{lightgreen}76.0 \\
        \cmidrule{2-8}
        & \multirow{3}{*}{SAA}  & ECAPA &  \cellcolor{lightestgreen}27.1 & \cellcolor{lightestgreen}49.0 & \cellcolor{lightergreen}62.2 & \cellcolor{lightergreen}68.3 & \cellcolor{lightgreen}72.1 \\
        & & MMS &  \cellcolor{lightergreen}51.4 & \cellcolor{midgreen1}85.4 & \cellcolor{midgreen2}91.0 & \cellcolor{midgreen2}90.4 & \cellcolor{midgreen1}87.0 \\
        & & GEO &  \cellcolor{midgreen1}86.6 & \cellcolor{darkergreen}\bf{97.2} & \cellcolor{darkergreen}96.2 & \cellcolor{darkgreen}93.4 & \cellcolor{midgreen1}88.5 \\
        \bottomrule
    \end{tabular}
    }
    \label{tab:accuracy_window_sizes_colored}
    \vspace{-3mm}
\end{table}

\section{Explaining accent-language confusion}
\label{sec:explaining_acc_lang_confusion}

One explanation for why LID models fail on accented speech is that they model differences in accent rather than differences between languages. 
While languages can be characterized by long-range lexical features, accents may be characterized by much shorter phonotactic features, such as the usage of certain phones or phone-grams.
Given that L2-accented speech often uses L1 phonotactics imposed over L2 words, models that only encode short or local features, rather than long-range lexical features, are likely to confuse such speech with L1 language speech. 
Thus, our hypothesis is that current LID models act as accent classifiers rather than language classifiers, explaining the observed pattern of accent-language confusion.

In order to explore this hypothesis, we examine how LID models capture long-range dependencies and their impact on accented speech classification.

\subsection{Block Permutation Invariance}
Models invariant to short block permutations may struggle to distinguish words with identical phonemes in different orders. To test this, we split 10s+ audio into $T$-second chunks, reverse their order, and analyze performance degradation.

Table \ref{tab:rev_results} shows performance of most models remains stable for chunk sizes down to 0-0.25s, roughly a phoneme’s duration. However, GEO degrades immediately on accented speech, indicating sensitivity to 0.5-2s sequences, i.e., the duration of 1-4 words. ECAPA-TDNN and MMS show minimal degradation, suggesting limited modeling of longer sequences.

\subsection{Long Range Dependency}
A model may be invariant to small block permutations yet still capture long-range dependencies, such as identifying a language through distant phoneme co-occurrences. To test whether models just aggregate local predictions or are capable of modeling long-range dependencies, we evaluate ECAPA-TDNN, MMS, and GEO on speech segments over 10 seconds from the SAA and EdAcc datasets. Segments are split into non-overlapping $T$-second chunks, and language predictions are aggregated by majority vote. Performance variation with chunk size indicates long-range modeling.

Table \ref{tab:accuracy_window_sizes_colored} confirms expected degradation on non-US english accents. However, ECAPA-TDNN improves with larger chunks, suggesting it \emph{does} capture long dependencies. Surprisingly, GEO improves (~10\% relative) on accented speech when context is \emph{limited}, indicating over-fitting to long sequences from common L1 accents. ECAPA-TDNN models short-term dependencies well, while GEO excels at longer sequence modeling. Notably, GEO—the least permutation-invariant model—proved most robust to L2 accents.

\section{Results and Discussion}

\label{sec:results}

See our results in \autoref{tab:approach_results}. We find that \etpstrain{} shows considerable gains on all accented speech over \et{} while maintaining a comparable performance on standard LID.\textsuperscript{3,4}\footnotetext[3]{We also obtained results for L1 minority accents in English (CommonVoice v1), German, French, Italian, and Spanish (CommonVoice v9.0) - e.g. Algerian French, Mexican Spanish. \etpstrain{} improves on average over \et{} on these data, and is consistently better on L1 accented speech for all languages except Italian.}\footnotetext[4]{On accented speech in Arabic, Spanish, and Chinese telephony from the NIST-LRE dataset \cite{sadjadi20182017}, performance was poor across all models and the acoustic domain mismatch appeared to dominate any other behaviors.} This, in conjunction with our error profile analysis above, validates our hypothesis that sequence-level information is beneficial in countering accent-language confusion for LID.

In fact, even simple fusion (\etps{}) maintains performance on FLEURS, and improves over \et{} on accented speech, presumably because the complementary error profiles of the two component models serve to amplify the correct vote and mute wrong answers.
\etpstrain{} yields further benefits by giving the model access to the \et{} representations during training, therefore allowing the model to learn a suitable combination strategy.

Note that \phoneseq{} by itself shows poor general performance, showing that acoustic representations are very informative for the task, even though they display accent-language confusion.
This is intuitive: in many cases, accent and language are indeed highly correlated, and short phonotactic features are very useful for LID on mainstream accented speech where the accent-language association may hold (unlike for L2-accent speech).

We observe that \etdutrain{} and \etduembedtrain{} show some improvements on accented speech but lag behind \etpstrain{}, indicating that explicit phone information is more useful than cluster centroid sequences. 
This may be because of the discrete units themselves encode some accent bias; it may also be a result of lossiness in the representations of input speech or the pooling / clustering steps.

\section{Conclusion}

Accented speech continues to pose a challenge to widely used LID systems today.
In this work, we characterize a systematic mode of error in SOTA LID systems for accented speech whereby L1-influenced L2-accented speech is classified as the L1 or a related language.
Our experiments, error profiling, permutation and context analyses, and ablations provide evidence that accent-language confusion is a major problem for mainstream LID models.
This behavior results from the failure to model long-enough input features and sequence order to characterize language rather than accent. 
We observe that models that are capable of modeling sequence information are accordingly more accent-robust.
Following the above insights, we show that explicitly incorporating sequence level information at the phoneme level mitigates accent-language confusion, resulting in significantly improved performance on L2-accented speech. 

\section{Acknowledgments}

We would like to thank Henry Li Xinyuan and Sanjeev Khudanpur for the helpful discussions.

\bibliographystyle{template}
\bibliography{template}

\end{document}